\documentclass{article}
\usepackage[T1]{fontenc}
\usepackage[preprint]{corl_2026}
\usepackage{microtype}
\usepackage{graphicx}
\usepackage{booktabs}
\usepackage{algorithm,algorithmic}

\usepackage{amsmath,amssymb,mathtools,amsthm}
\makeatletter
\@ifundefined{lemma}{}{}
\@ifundefined{assumption}{}{}
\@ifundefined{definition}{}{}
\@ifundefined{proposition}{}{}
\@ifundefined{corollary}{}{}
\theoremstyle{remark}
\@ifundefined{remark}{}{}
\theoremstyle{plain}
\makeatother
\usepackage[capitalize,noabbrev]{cleveref}
\usepackage{xcolor}
\DeclareRobustCommand{\echomethod}{\textcolor[HTML]{007F86}{\textbf{Echo}}}
\DeclareRobustCommand{\sdpgmethod}{\textcolor[HTML]{64748B}{\textbf{SDPG}}}
\definecolor{echolink}{HTML}{007F86}
\hypersetup{colorlinks=true,linkcolor=echolink,citecolor=echolink,urlcolor=echolink}

\title{Echo: Scaling Visual Reinforcement Learning with  Off-Policy Experience}
\author{
  Gagan Khandate\\
  \texttt{gagank@cs.columbia.edu}
}
\hypersetup{pdftitle={ECHO: Scaling Visual Reinforcement Learning with Off-Policy  Experience},pdfauthor={Gagan Khandate}}
\begin{document}
\maketitle

\begin{abstract}
Scaling visual reinforcement learning (RL) with GPU-accelerated simulation is computationally expensive due to rendering and visual-policy inference costs. Consequently, two-stage teacher--student distillation is commonly used to train visual policies, but introduces an imitation gap. In this preliminary study, we introduce \echomethod{}, \textbf{a single-stage off-policy RL framework that scales experience collection without proportionally increasing visual computation}. Through \textbf{inverse distillation}, a state-based auxiliary policy imitates the evolving visual policy and collects inexpensive, unrendered experience. Echo leverages this experience for off-policy value learning, directly improving the visual policy through a shared privileged critic. We instantiate \echomethod{} with FastTD3 and evaluate it on three visual control tasks and two dexterous manipulation tasks in Genesis. Relative to \sdpgmethod{} under the evaluated configurations, Echo reduces time to reference returns by an average of \textbf{60\% across the three locomotion tasks} and \textbf{56\% across both hand tasks}, while improving peak mean returns by \textbf{62\% and 12\%}, respectively. Both hand tasks also achieve higher mean returns at matched simulator-transition budgets. These initial results suggest that scaling off-policy experience is a promising approach to efficient direct visual RL.
\end{abstract}


\begin{figure}[htbp]
\centering\includegraphics[width=\textwidth]{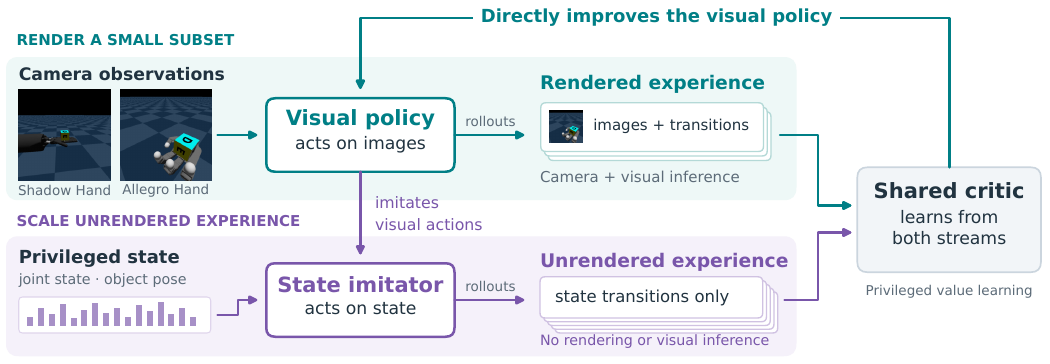}
\caption{\textbf{Scaling visual RL with off-policy auxiliary experience.} A state-based imitator collects unrendered transitions, while a small rendered subset supplies visual-policy experience. Both streams train a privileged critic that directly improves the visual policy.}

\label{fig:echo-overview}
\end{figure}

\clearpage
\section{Introduction}

Massively parallel simulation has expanded the scale of reinforcement learning (RL) and underpins many recent advances in robot learning, including simulated driving \citep{kazemkhani2025gpudrive}, locomotion \citep{rudin2022walk,zhang2026sgs}, and contact-rich assembly \citep{zhang2026alphanist}. While parallel physics simulation enables collecting large volumes of experience, visual control introduces additional computational costs from observation \textbf{rendering and visual-policy inference}, which can grow substantially with model size. Consequently, scaling visual RL remains considerably more expensive than scaling state-based RL.

Teacher--student distillation is a common approach to training visual control policies using privileged simulator state \citep{miki2022perceptive,wang2024cts,srinivasan2025acgd}. A state-based teacher first learns the task and subsequently supervises a visual student through its actions. However, the teacher may condition its behavior on information unavailable to the student, making certain decisions difficult to reproduce \citep{kim2025realizable}. This information asymmetry creates an \emph{imitation gap}, further compounded by distribution shift resulting from student errors \citep{ross2011dagger}. Directly optimizing the visual policy avoids this teacher--student mismatch, but reintroduces the computational costs of visual experience collection.

Recent works have explored more efficient approaches to direct visual RL. Squint combines parallel simulation with low-resolution visual observations \citep{almuzairee2026squint}, although aggressive downsampling may discard visual details essential for precise manipulation. More recently, \sdpgmethod{} reduces rendering costs by estimating policy improvements from perturbed, unrendered rollouts using zeroth-order optimization \citep{you2026sdpg}. While zeroth-order methods avoid differentiating through contact dynamics, their update directions are estimated from finite samples of perturbed trajectories. In contact-rich manipulation, small perturbations can induce different contact modes and substantially different returns, potentially increasing estimator variance. Critic-based first-order methods offer an alternative by providing learned, differentiable optimization signals, although their effectiveness depends on the accuracy of the value function. Furthermore, the on-policy formulation of \sdpgmethod{} limits the reuse of previously collected experience.

Recent advances in off-policy RL have demonstrated large scale off-policy experience can be effectively combined with massively parallel simulation. Methods such as FastTD3 \citep{seo2025fasttd3} and FlashSAC \citep{kim2026flashsac} leverage parallel experience collection, replay-based value learning, and efficient critic optimization to achieve rapid learning in high-dimensional robotic control tasks. These advances demonstrate the potential of off-policy RL to exploit large volumes of simulation experience through repeated value and policy updates. However, directly extending this paradigm to visual RL remains computationally expensive, as collecting additional experience using a visual policy requires rendering observations and evaluating the policy at every environment step. This raises a central question: \emph{can we scale off-policy experience collection for visual RL without proportionally scaling rendering and visual-policy inference?}

We present a preliminary study of \echomethod{}, a \textbf{single-stage off-policy RL framework} that addresses this challenge through \textbf{inverse distillation}. Unlike conventional teacher--student distillation, where a privileged state-based teacher supervises a visual student, Echo reverses the direction of supervision: an auxiliary state-based policy learns to \textbf{imitate the evolving visual policy}. This state-based imitator enables inexpensive experience collection \textbf{without camera rendering or visual-policy inference}, while approximately tracking the state distribution induced by the visual policy. Crucially, the additional unrendered experience is not merely used to supervise the visual policy, but directly contributes to its improvement through off-policy reinforcement learning.

Specifically, Echo learns a \textbf{shared privileged critic} from both \textbf{rendered and unrendered experience} stored in replay. The state-based imitator serves two complementary purposes: (1) collecting inexpensive transitions that remain approximately in-distribution with the evolving visual policy, and (2) providing \textbf{continuation actions} for temporal-difference updates of the state-based critic. The visual policy is then improved directly through \textbf{critic-guided first-order updates} using a smaller set of paired state--observation samples. By separating large-scale experience collection from expensive visual-policy evaluation, Echo combines the scalability and experience reuse of off-policy RL with the benefits of direct visual policy optimization, without requiring a separately trained privileged teacher.

Our contributions are:
\begin{itemize}
    \item \textbf{Scalable off-policy visual RL.} We introduce \echomethod{}, a framework that decouples the scale of experience collection from visual rendering and inference, enabling large-scale off-policy value learning from inexpensive, unrendered simulation transitions.
    \item \textbf{Inverse distillation for experience collection.} We propose a state-based imitator trained to track the evolving visual policy, reversing conventional teacher--student supervision. The imitator enables approximately on-distribution auxiliary rollouts and supplies continuation actions for off-policy critic learning, allowing unrendered experience to directly improve the visual policy.
    \item \textbf{Preliminary evaluation on visual control and dexterous manipulation.} We instantiate \echomethod{} with FastTD3 and evaluate it on three visual control tasks and two dexterous manipulation tasks in Genesis. Relative to \sdpgmethod{} under the evaluated configurations, Echo reduces time to reference returns by an average of \textbf{60\% across the three locomotion tasks} and \textbf{56\% across both hand tasks}, with \textbf{62\% and 12\% higher peak mean returns}, respectively. Both hands also achieve higher mean returns at matched simulator-transition budgets. Appendix~\ref{app:aggregate-performance} summarizes these descriptive averages and the Ant baseline-selection and compute qualifications.
\end{itemize}

\section{Related Work}
\label{sec:related}

\paragraph{Privileged learning and policy distillation.}
Simulation provides privileged state information that can facilitate visual policy learning without being available at deployment. Asymmetric actor--critic methods use privileged state for value estimation while restricting the actor to deployment observations \citep{pinto2017asymmetric}. Alternatively, teacher--student distillation trains a privileged state-based policy and transfers its behavior to a visual student \citep{levine2016visuomotor,miki2022perceptive}. More recent approaches train the teacher and student concurrently, combining action supervision with direct policy improvement \citep{wang2024cts,srinivasan2025acgd}. However, privileged teachers may rely on information unavailable to the visual student, creating an imitation gap \citep{kim2025realizable}. Related approaches have also leveraged privileged policies for experience collection. Schwab et al. employ feature-based auxiliary policies to generate experience for vision-based control \citep{schwab2019simultaneously}, while APRiL trains state-based and visual agents with shared experience \citep{salter2021april}. In contrast, \echomethod{} employs \emph{inverse distillation}, where a state-based auxiliary policy imitates the evolving visual policy to collect inexpensive experience for off-policy value learning, rather than supervising the visual policy.

\paragraph{Scaling off-policy reinforcement learning.}
Massively parallel GPU simulation has enabled rapid reinforcement learning for robotic control \citep{makoviychuk2021isaacgym,rudin2022walk}. Recent methods such as FastTD3 \citep{seo2025fasttd3} and FlashSAC \citep{kim2026flashsac} extend these scaling benefits to off-policy RL, leveraging parallel experience collection, replay buffers, and efficient critic optimization. By reusing collected transitions across optimization updates, off-policy methods can improve sample efficiency while benefiting from high simulation throughput. However, applying this paradigm directly to visual RL incurs substantial rendering and visual-policy inference costs as experience collection scales. \echomethod{} addresses this bottleneck by collecting additional state-based experience for off-policy critic learning without requiring visual observations at every simulation step.

\paragraph{Efficient visual reinforcement learning.}
Recent methods have reduced the computational costs of direct visual RL. Squint uses low-resolution visual observations with massively parallel simulation to reduce rendering and inference costs \citep{almuzairee2026squint}. \sdpgmethod{} limits rendering to a subset of environments and estimates policy improvements from unrendered perturbation rollouts using zeroth-order optimization \citep{you2026sdpg}. While these approaches reduce visual computation, they do not exploit large-scale unrendered experience through off-policy value learning. \echomethod{} instead trains a shared privileged critic using both rendered and unrendered transitions, with a state-based imitator providing auxiliary experience and approximate continuation actions for temporal-difference learning. This enables direct visual policy improvement while decoupling the scale of off-policy learning from rendering and visual-policy inference costs.

\section{Problem Formulation}
\label{sec:preliminaries}

We formulate control as a Markov decision process $\mathcal M=(\mathcal S,\mathcal A,P,R,\rho_0,\gamma)$, where $\mathcal S$ is the state space, $\mathcal A$ is a continuous action space, $P$ describes the transition dynamics, $R$ defines the reward, $\rho_0$ is the initial-state distribution, and $\gamma\in[0,1)$ is the discount factor.

We distinguish between two sources of information available in simulation: privileged state $s_t\in\mathcal S$, which can be accessed inexpensively during training, and observations $o_t$, which may require costly rendering or sensor simulation. The latter include camera images and other sensory inputs, together with permitted proprioception, and constitute the only inputs available to the deployed policy. We train a deterministic, feedforward policy $\pi_\theta(o_t)$ to maximize
\begin{equation}
J(\theta)=\mathbb E_{\rho_0,P,\pi_\theta}\left[\sum_{t=0}^{\infty}\gamma^t R(s_t,\pi_\theta(o_t))\right].
\label{eq:control-objective}
\end{equation}

During training, the simulator additionally exposes privileged state $s_t$. We use this state in an auxiliary policy $\mu_\psi(s)$ and a differentiable state--action critic $Q_\phi(s,a)$; only the visual policy $\pi_\theta$ is deployed. The notation $\operatorname{sg}[\cdot]$ denotes a stopped gradient, and bars denote slowly updated target parameters.

\begin{figure}[t]
\centering\includegraphics[width=\textwidth]{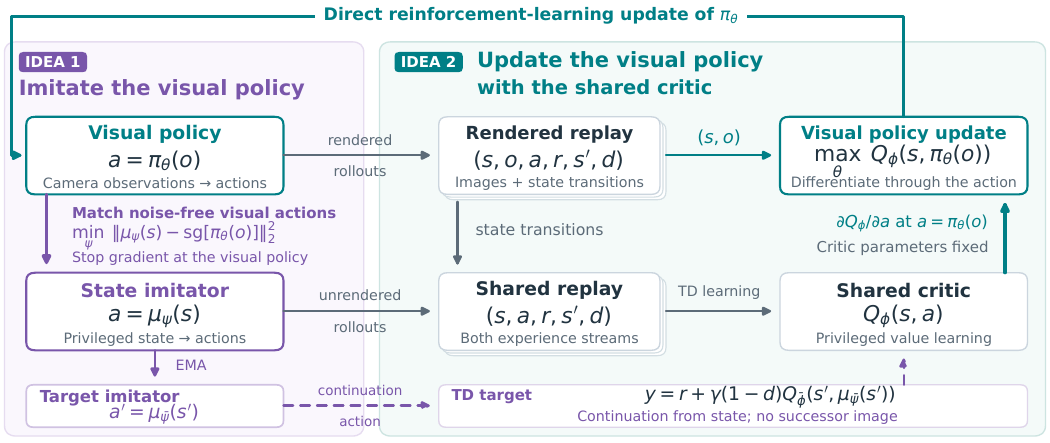}

\caption{\textbf{Two key ideas in \echomethod{}.} \textcolor[HTML]{7956AA}{\textbf{(1)}} Imitate the visual policy: a state-based imitator matches noise-free visual actions with stopped gradients. \textcolor[HTML]{007F86}{\textbf{(2)}} Update the visual policy with the shared critic: both experience streams train the critic, whose action gradient improves the visual policy while critic parameters remain fixed. The imitator's EMA copy supplies TD continuation actions without successor images (dashed purple). Gray arrows denote experience flow.}
\label{fig:echo-method}
\end{figure}

\section{Method}
\label{sec:echo}

\echomethod{} combines inverse policy distillation with off-policy actor--critic learning to scale visual reinforcement learning without proportionally increasing rendering or visual-policy inference costs. A state-based auxiliary policy imitates the evolving visual policy and collects additional experience in unrendered environments. A shared privileged critic learns from both rendered and unrendered transitions, providing action gradients to improve the visual policy directly. Together, these components decouple the scale of experience collection and value learning from the computational cost of visual-policy evaluation, without requiring a privileged teacher to solve the task before visual policy learning begins. The following sections describes each of these components.
Figures~\ref{fig:echo-overview} and~\ref{fig:echo-method} illustrate the framework; Algorithm~\ref{alg:echo} summarizes the training procedure.

\begin{algorithm}[t]
\caption{\echomethod{}}
\label{alg:echo}
\begin{algorithmic}[1]
\STATE Initialize visual policy $\pi_\theta$, state-based imitator $\mu_\psi$, and twin distributional critics $Z_{\phi_k}$, $k\in\{1,2\}$; write $\phi=(\phi_1,\phi_2)$.
\STATE Set $\bar\psi=\psi$, $\bar\phi_k=\phi_k$, and initialize replay buffers $\mathcal D$ and $\mathcal D_{\mathrm{nom}}$.
\FOR{each training iteration}
    \STATE Collect experience in $M$ rendered environments using $\pi_\theta(o)$ and $N-M$ unrendered environments using the online imitator $\mu_\psi(s)$, with exploration noise and clipped actions.
    \STATE Store $(s,a,r,s',d)$ from both streams in $\mathcal D$; store rendered $(s,o)$ pairs in $\mathcal D_{\mathrm{nom}}$.
    \STATE Form $B_{\mathrm{aux}}$ from fresh rendered pairs $(s,o)\sim\rho_{\mathrm{nom}}$ and noise-free actions $\pi_\theta(o)$.
    \STATE Estimate $\widehat{\mathcal L}_{\mathrm{aux}}(\psi)$ on $B_{\mathrm{aux}}$ using Equation~\eqref{eq:aux-imitation}; update $\psi\leftarrow\psi-\eta_\mu\nabla_\psi\widehat{\mathcal L}_{\mathrm{aux}}(\psi)$.
    \STATE $\bar\psi\leftarrow(1-\tau_\mu)\bar\psi+\tau_\mu\psi$.
    \STATE Sample transition minibatch $B_Q\sim\mathcal D$.
    \STATE Update both $Z_{\phi_k}$ on $B_Q$ using projected categorical Bellman targets from $Z_{\bar\phi_k}$ and continuation policy $\mu_{\bar\psi}$, with clipped double-Q learning and target-policy smoothing. Equation~\eqref{eq:critic-objective} gives the scalar analogue.
    \STATE Sample paired minibatch $B_\pi\sim\mathcal D_{\mathrm{nom}}$.
    \STATE Estimate $\widehat{\mathcal J}_{\mathrm{vis}}(\theta;\phi)$ on $B_\pi$ using Equation~\eqref{eq:actor-objective}; update $\theta\leftarrow\theta+\eta_\pi\nabla_\theta\widehat{\mathcal J}_{\mathrm{vis}}(\theta;\phi)$, holding $\phi$ fixed.
    \STATE $\bar\phi_k\leftarrow(1-\tau_Q)\bar\phi_k+\tau_Q\phi_k$ for $k\in\{1,2\}$.
\ENDFOR
\STATE Deploy $\pi_\theta(o)$ using deployment observations only.
\end{algorithmic}
\end{algorithm}

\subsection{Inverse Distillation for Rendering-Free Experience Collection}
\label{sec:echo-imitation}

The first component of \echomethod{} is a state-based auxiliary policy that follows the evolving visual policy, rather than serving as a privileged teacher. By approximating visual-policy actions from simulator state, this auxiliary policy can collect additional experience without rendering observations or evaluating the visual network.

\begingroup
\setlength{\fboxsep}{6pt}
\setlength{\fboxrule}{0.5pt}
\noindent\fcolorbox[HTML]{D0BFE6}{F5F0FA}{%
\begin{minipage}{\dimexpr\linewidth-2\fboxsep-2\fboxrule\relax}
\textbf{\textcolor[HTML]{7956AA}{Idea 1}} Distill the evolving visual policy into a state-based auxiliary policy that collects unrendered experience.
\end{minipage}}
\endgroup
\par\medskip

\paragraph{Rendered and Unrendered Experience.}
We partition $N$ parallel environments into two streams. In $M$ rendered (nominal) environments, the visual policy $\pi_\theta(o)$ acts using visual observations, and we retain paired states and observations. In the remaining $N-M$ environments, the auxiliary policy $\mu_\psi(s)$ acts directly from privileged simulator state, avoiding observation rendering and visual-policy inference. Both streams use exploration noise, with actions clipped to the task's action bounds.

A shared replay buffer $\mathcal D$ stores transitions $(s,a,r,s',d)$ from both streams, where $s'$ denotes the successor state before any reset and $d$ indicates true termination. Time-limit truncations retain bootstrapping. A separate rendered replay buffer $\mathcal D_{\mathrm{nom}}$ stores paired state--observation samples $(s,o)$ for visual-policy optimization. Thus, all collected experience contributes to critic learning, while updates requiring visual observations are restricted to rendered samples.

\paragraph{Inverse Policy Distillation.}
The auxiliary policy learns to reproduce the current visual policy's noise-free actions using privileged simulator state. Fresh rendered state--observation pairs define the nominal distribution $\rho_{\mathrm{nom}}$, yielding the imitation objective

\begin{equation}
\mathcal L_{\mathrm{aux}}(\psi)=
\mathbb E_{(s,o)\sim\rho_{\mathrm{nom}}}
\left[
\|\mu_\psi(s)-\operatorname{sg}[\pi_\theta(o)]\|_2^2
\right].
\label{eq:aux-imitation}
\end{equation}

Here, $\operatorname{sg}$ denotes stop-gradient, ensuring that imitation updates only the auxiliary policy. As the visual policy improves, the auxiliary policy continually tracks its behavior, providing a lightweight state-based approximation that can be evaluated without rendering. The online auxiliary policy collects additional experience, while an exponential moving average (EMA) copy supplies continuation actions for critic learning.

\subsection{Shared Critic Learning and Visual Policy Improvement}
\label{sec:echo-shared-critic}

The second component of \echomethod{} is off-policy value learning using a shared privileged critic trained on experience collected by both the visual policy and its state-based imitator. The critic enables the visual policy to benefit from large-scale unrendered experience through first-order policy improvement, without requiring visual observations for every transition.

\begingroup
\setlength{\fboxsep}{6pt}
\setlength{\fboxrule}{0.5pt}
\noindent\fcolorbox[HTML]{9ACEC6}{EDF8F5}{%
\begin{minipage}{\dimexpr\linewidth-2\fboxsep-2\fboxrule\relax}
\textbf{\textcolor[HTML]{007F86}{Idea 2}} Scale off-policy value learning with rendered and unrendered experience, and improve the visual policy directly through critic gradients.
\end{minipage}}
\endgroup
\par\medskip

\paragraph{Learning the Shared Critic without Successor Images.}
We train the shared critic using transitions sampled from both rendered and unrendered replay. Although unrendered transitions lack visual observations, their privileged successor states remain available. The state-based imitator provides approximate continuation actions for the visual policy, allowing us to construct temporal-difference (TD) targets without rendering successor observations or evaluating the visual policy.

For clarity, we first describe the learning objective using a scalar action-value function $Q_\phi(s,a)$. Given a transition $(s,a,r,s',d)$ sampled from the combined replay distribution $\mathcal D$, the one-step TD target and critic objective are

\begin{equation}
\begin{aligned}
y &= r+\gamma(1-d)
Q_{\bar\phi}(s',\mu_{\bar\psi}(s')),\\
\mathcal L_Q(\phi)&=
\mathbb E_{\mathcal D}
\left[
\left(Q_\phi(s,a)-\operatorname{sg}[y]\right)^2
\right],
\end{aligned}
\label{eq:critic-objective}
\end{equation}
where $\mu_{\bar\psi}$ and $Q_{\bar\phi}$ denote target networks and $\operatorname{sg}[\cdot]$ stops gradients.

The target networks are updated using exponential moving averages:
\begin{equation}
\begin{aligned}
\bar\psi &\leftarrow (1-\tau_\mu)\bar\psi+\tau_\mu\psi,\\
\bar\phi &\leftarrow (1-\tau_Q)\bar\phi+\tau_Q\phi,
\end{aligned}
\end{equation}
where $\tau_\mu,\tau_Q\in(0,1]$. These updates smooth changes in continuation actions and TD targets as the visual policy and its imitator evolve.

\paragraph{Distributional Critic Implementation.}
In practice, we instantiate the shared critic using FastTD3 \citep{seo2025fasttd3}, which employs twin distributional critics rather than a scalar Q-function. Each critic $Z_{\phi_k}(s,a)$ predicts a categorical distribution over discounted returns, with expected action value
\begin{equation}
Q_{\phi_k}(s,a)
=
\sum_{i=1}^{N_Z}
z_i p_{\phi_k,i}(s,a),
\label{eq:distributional-q}
\end{equation}
where $\{z_i\}_{i=1}^{N_Z}$ is a fixed support of return atoms and $p_{\phi_k,i}(s,a)$ denotes the predicted probability of atom $z_i$.

The scalar TD objective in Eq.~\eqref{eq:critic-objective} illustrates the role of the auxiliary continuation policy. In our implementation, the critics instead minimize a categorical cross-entropy loss against projected distributional Bellman targets, with clipped double-Q learning and target-policy smoothing as in FastTD3. Importantly, both critics learn from rendered and unrendered experience. Further implementation details are provided in Appendix~\ref{app:implementation}.

\paragraph{First-Order Visual Policy Improvement.}
The visual policy is optimized directly using paired privileged states and rendered observations from nominal replay $\mathcal D_{\mathrm{nom}}$. Holding the critic parameters fixed, we maximize the clipped expected action value of the twin distributional critics:

\begin{equation}
\mathcal J_{\mathrm{vis}}(\theta;\phi)=
\mathbb E_{(s,o)\sim\mathcal D_{\mathrm{nom}}}
\left[
Q_\phi(s,\pi_\theta(o))
\right].
\label{eq:actor-objective}
\end{equation}

Here, the scalar objective used for visual policy improvement is
\begin{equation}
Q_\phi(s,a)
=
\min_{k\in\{1,2\}}Q_{\phi_k}(s,a).
\label{eq:clipped-actor-q}
\end{equation}

The visual policy recomputes its actions from replayed observations. Gradients propagate through the critic's expected action value into the visual policy:
\begin{equation}
\nabla_\theta \mathcal J_{\mathrm{vis}}
=
\mathbb E_{(s,o)\sim\mathcal D_{\mathrm{nom}}}
\left[
\nabla_\theta\pi_\theta(o)
\nabla_a Q_\phi(s,a)
\big|_{a=\pi_\theta(o)}
\right].
\label{eq:visual-policy-gradient}
\end{equation}

Unlike zeroth-order methods that estimate policy improvements through sampled trajectory perturbations, Echo uses first-order gradients of a learned action-value function. The shared critic reuses experience from both policies to estimate action values, providing a differentiable optimization signal for the visual actor. While this avoids repeatedly estimating update directions from perturbed trajectories, the quality of the resulting gradients depends on the accuracy of the learned critic.

Crucially, unrendered experience contributes to visual-policy improvement through the shared critic, without requiring rendering or visual-policy inference for those transitions. Only the smaller set of rendered state--observation pairs is needed to propagate gradients through the visual encoder and actor. Appendix~\ref{app:critic-interpretation} discusses the interpretation of these gradients and the approximation introduced by auxiliary continuation actions.

\section{Experimental Results}
\label{sec:practical}

\begin{figure}[t]
\centering
\includegraphics[width=\textwidth]{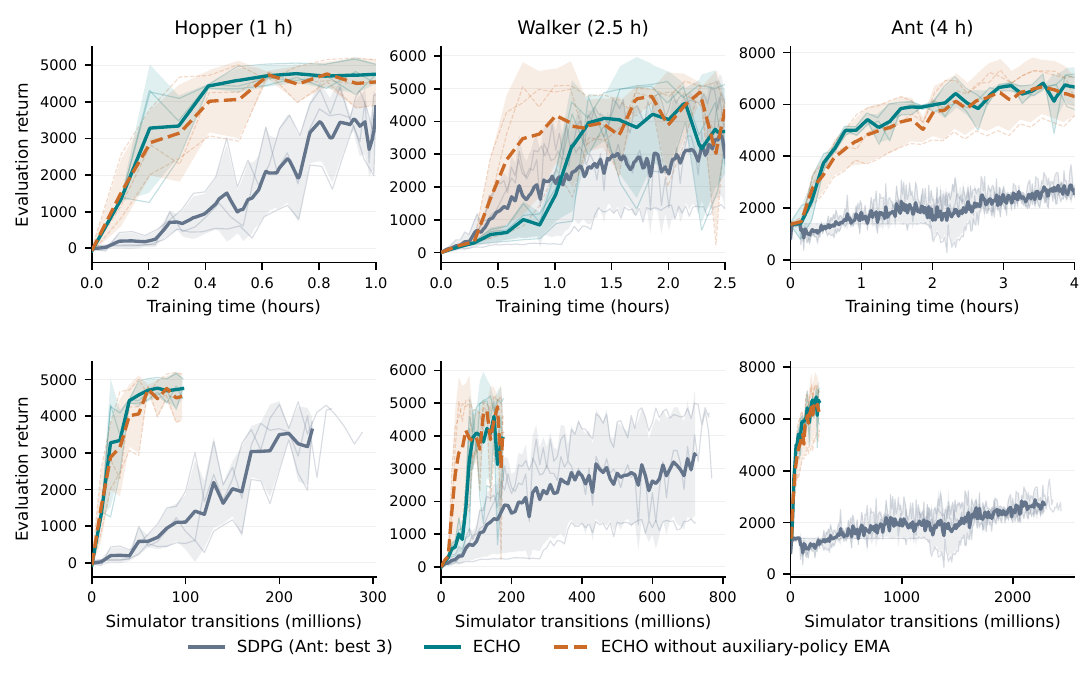}
\caption{\textbf{Visual locomotion benchmarks in Genesis.} Episode returns as a function of wall-clock training time (top, excluding evaluation) and total simulator transitions (bottom, including auxiliary interactions). Solid lines and shaded bands indicate mean and sample standard deviation across three seeds, respectively; faint lines show individual runs. Reference hyperparameters are in Appendix~\ref{app:implementation}; the Ant baseline qualification is in Appendix~\ref{app:aggregate-performance}.}
\label{fig:locomotion-results}
\end{figure}

\begin{figure}[t]
\centering
\includegraphics[width=0.8\textwidth]{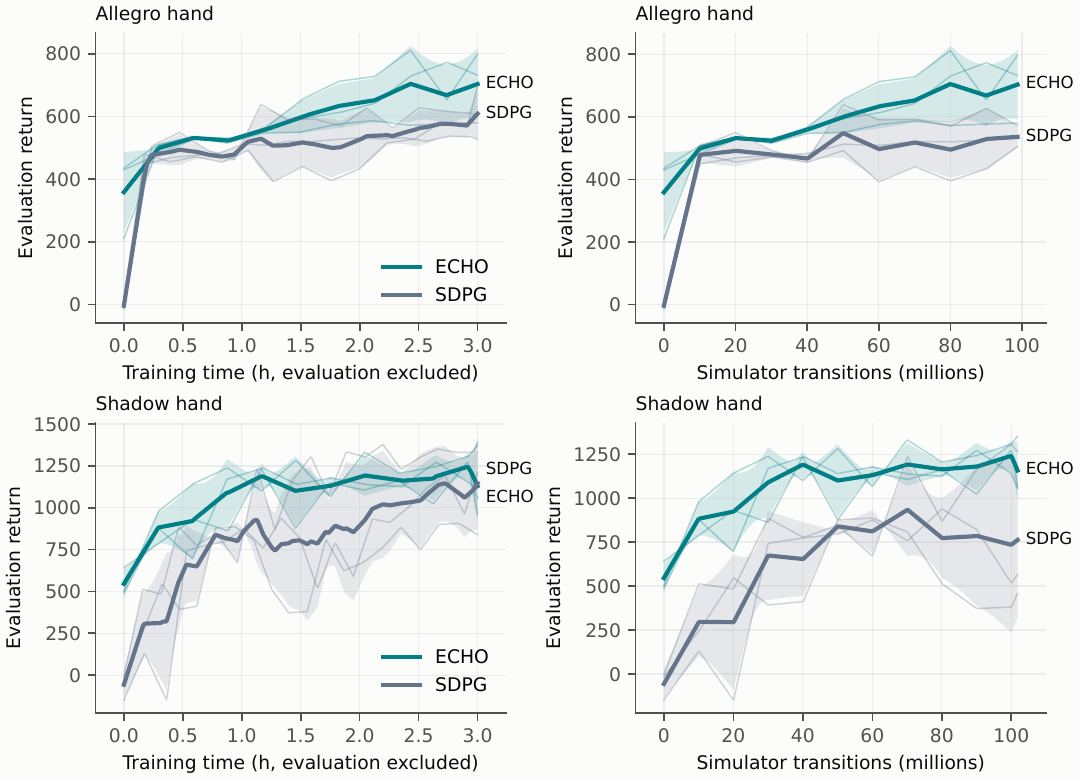}
\caption{\textbf{Visual dexterous manipulation in Genesis.} Episode returns on Allegro Hand (top) and Shadow Hand (bottom) as a function of wall-clock training time (left, excluding evaluation) and total simulator transitions (right, including auxiliary interactions). All runs use a three-hour training budget. Solid lines and shaded bands indicate mean and sample standard deviation across three seeds, respectively; faint lines show individual runs. Curves are linearly interpolated over the common support of all six runs. Differences in initial returns arise from shaped rewards and policy initialization.}
\label{fig:manipulation}
\end{figure}

\begin{figure}[t]
\centering
\includegraphics[width=\textwidth]{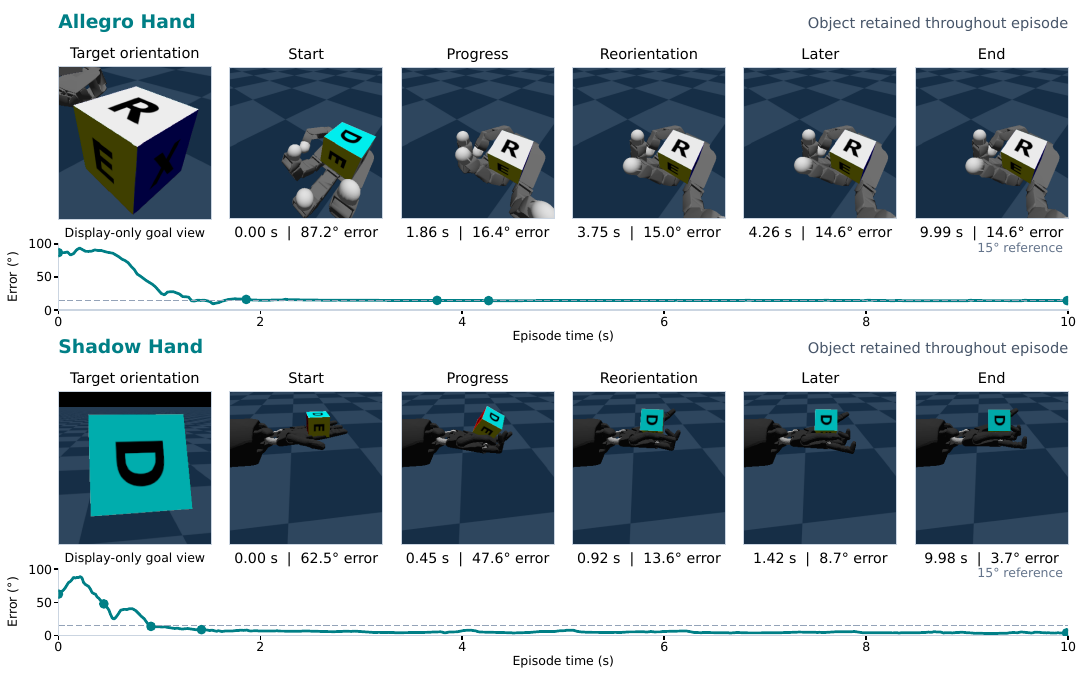}\caption{\textbf{\echomethod{} dexterous manipulation rollouts.} Selected in-hand object reorientation episodes on Allegro Hand and Shadow Hand, showing target orientations, timestamps, and measured orientation errors. Both policies maintain object grasp throughout the episodes. Rollouts are generated using earlier checkpoints than those evaluated in Figure~\ref{fig:manipulation} and are illustrative rather than representative of aggregate success rates.}
\label{fig:hand-rollouts}
\end{figure}

\begin{figure}[t]
\centering
\includegraphics[width=\textwidth]{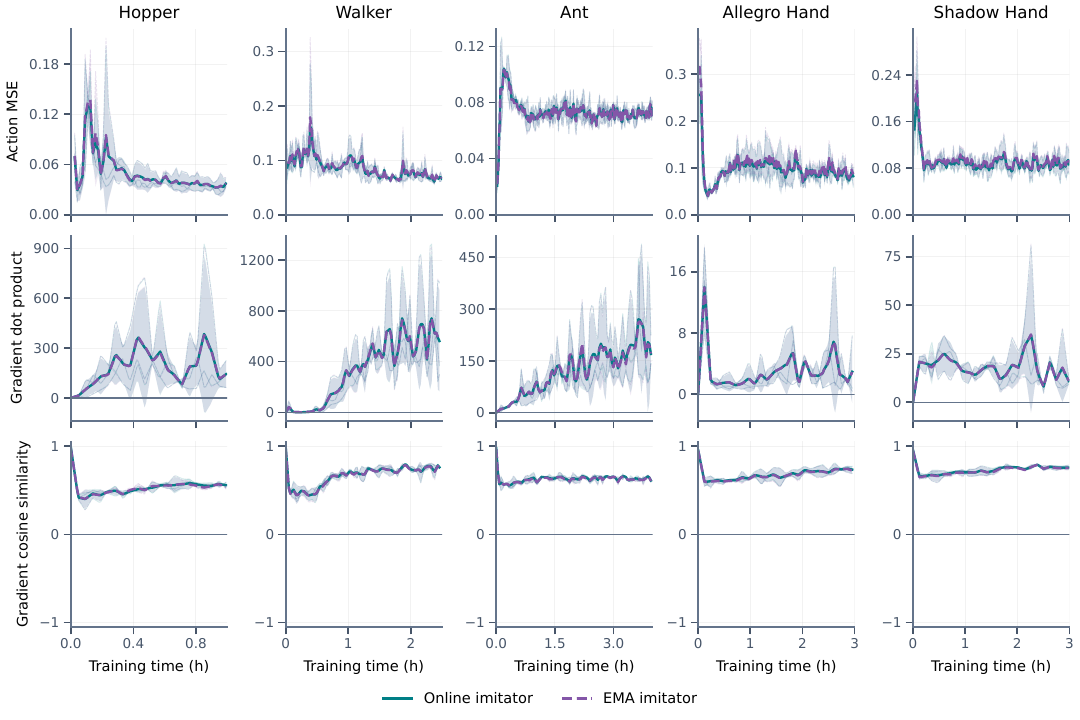}
\caption{\textbf{Inverse distillation and critic-gradient diagnostics.} (Top) Online imitation loss (teal, solid) and post-update EMA action error (purple, dashed) on fresh visual-policy states, averaged over 60-second intervals. (Middle and bottom) Critic action-gradient dot product and cosine similarity between visual-policy and imitator actions on paired rendered replay states, measured in separate diagnostic runs. Lines and shaded bands show mean and sample standard deviation across three seeds; faint lines show individual runs.}
\label{fig:diagnostics}
\end{figure}

Our experiments investigate whether scaling off-policy experience through inverse distillation improves the efficiency of direct visual reinforcement learning. We evaluate \echomethod{} on visual control and dexterous manipulation tasks in Genesis, addressing three questions:

\begin{enumerate}
    \item \textbf{Learning efficiency:} How does \echomethod{} compare with \sdpgmethod{} in wall-clock training time, simulator-transition efficiency, and achieved returns?
    \item \textbf{Auxiliary policy design:} How does exponential moving average (EMA) stabilization of the state-based imitator affect learning performance and variability across seeds?
    \item \textbf{Learning dynamics:} How accurately does the state-based imitator track the evolving visual policy, and how do their critic action gradients compare?
\end{enumerate}

\subsection{Compared Methods and Evaluation Protocol}

\paragraph{\echomethod{}.}
We instantiate \echomethod{} with FastTD3 \citep{seo2025fasttd3}, retaining its twin categorical critics, target-policy smoothing, and delayed actor updates. We extend FastTD3 with a state-based imitator trained through inverse distillation for auxiliary experience collection, with an exponential moving average (EMA) copy supplying continuation actions for critic learning.

\paragraph{\sdpgmethod{}.}
We evaluate \sdpgmethod{} using its publicly released implementation \citep{you2026sdpg}. Both methods reduce visual computation by limiting rendering to a subset of parallel environments, but differ in how they leverage unrendered experience. While \sdpgmethod{} uses perturbation-based policy improvement, \echomethod{} exploits auxiliary experience through off-policy value learning. Our experiments compare the complete implementations rather than matched architectures or hyperparameters.

\paragraph{Evaluation protocol.}
Unless otherwise specified, we evaluate each method over three training seeds (0, 1, and 2), using deterministic visual-policy evaluations over 64 episodes. We report mean returns and sample standard deviations across seeds. Wall-clock training time excludes evaluation, while simulator-transition counts include both rendered and auxiliary interactions. In the main sweep, three training processes share each GPU. Reference hyperparameters are provided in Appendix~\ref{app:implementation}; the Ant baseline qualification and aggregate comparisons are described in Appendix~\ref{app:aggregate-performance}.


\subsection{Tasks}
We evaluate three visual locomotion tasks (Hopper, Walker, and Ant) and two dexterous manipulation tasks (Allegro Hand and Shadow Hand) using the Genesis implementations released by \citet{you2026sdpg}. The locomotion tasks are adapted from the MuJoCo benchmark, while the dexterous manipulation tasks involve in-hand cube reorientation with egocentric visual observations.

Training budgets are one hour for Hopper, 2.5 hours for Walker, four hours for Ant, and three hours for each hand task. We report returns against both wall-clock training time and total simulator transitions, including auxiliary experience collected by \echomethod{}.

\subsection{Training Performance}

\paragraph{Visual locomotion.}
\echomethod{} achieves higher mean final returns and reaches comparable performance earlier than \sdpgmethod{} on the visual locomotion tasks (Figure~\ref{fig:locomotion-results}), subject to the qualification for Ant discussed in Appendix~\ref{app:aggregate-performance}. On Hopper, \echomethod{} reaches the \sdpgmethod{} reference return in approximately 21 minutes, compared with 60 minutes for \sdpgmethod{}, corresponding to a \textbf{2.9$\times$ improvement in wall-clock learning efficiency}. Walker exhibits a smaller timing advantage.

\paragraph{Dexterous manipulation.}
On Allegro Hand, \echomethod{} reaches the \sdpgmethod{} reference return in approximately 96 minutes, compared with 180 minutes for \sdpgmethod{}, corresponding to a \textbf{1.9$\times$ improvement in wall-clock learning efficiency}. It also achieves a higher final mean return of approximately 700, compared with 600 for \sdpgmethod{} (Figure~\ref{fig:manipulation}). On Shadow Hand, \echomethod{} reaches comparable performance earlier, while both methods achieve final mean returns near 1100. Thus, the principal advantage on Shadow Hand is faster learning rather than higher final performance.

\paragraph{Simulator-transition efficiency.}
The advantages extend beyond wall-clock training time. At matched simulator-transition counts, \echomethod{} achieves higher mean returns on both dexterous manipulation tasks, even when auxiliary interactions are included in the transition budget. On Hopper and Walker, \echomethod{} achieves higher returns while using approximately three to four times fewer simulator transitions at the assigned training budgets. These results suggest that large-scale auxiliary experience can improve the efficiency of direct visual policy learning. However, the comparisons involve complete implementations and do not isolate the effects of auxiliary experience from differences in backbone architectures and training configurations.

\subsection{Design Choices and Ablations}

\paragraph{Auxiliary-policy averaging.}
We examine the effect of exponential moving average (EMA) stabilization of the auxiliary policy by comparing $\tau_\mu=0.1$ against $\tau_\mu=1$, corresponding to no averaging. In both configurations, the online imitator collects auxiliary experience, while EMA changes the continuation actions used in critic targets. EMA produces task-dependent results (Figure~\ref{fig:locomotion-results}). On Hopper and Ant, it increases final mean returns by 210 and 369, respectively, and reduces across-seed variation, although the reduction on Ant is small. In contrast, removing EMA improves final mean return on Walker by 687 and reduces variation. These results indicate that EMA can benefit some tasks but does not consistently improve performance or stability.

\subsection{Learning Dynamics}

To better understand the interaction between inverse distillation and off-policy value learning, we analyze auxiliary-policy imitation error and critic action-gradient alignment (Figure~\ref{fig:diagnostics}).

\paragraph{Tracking the visual policy.}
The state-based imitator is trained to follow an evolving visual policy rather than a fixed behavioral target. After initial transients, imitation errors remain nonzero throughout training. Online and EMA imitation errors become similar later in training, with larger early differences on the dexterous manipulation tasks. These measurements show that auxiliary imitation remains imperfect while visual-policy learning progresses. However, imitation error is measured on states visited by the visual policy, and therefore does not establish accurate tracking on auxiliary-rollout states or unbiased temporal-difference targets.

\paragraph{Action-gradient alignment.}
We examine whether the shared critic produces consistent local action-improvement directions at actions selected by the visual policy and its state-based imitator. Using paired rendered replay samples, we compare critic action gradients across three seeds per task (Appendix~\ref{app:gradient-alignment}). Mean cosine similarity remains positive after the initial transient, with nearly identical trends for the online and EMA imitators. At the final common observation time, online/EMA cosine similarities are 0.56/0.55 on Hopper, 0.76/0.75 on Walker, 0.61/0.60 on Ant, 0.73/0.72 on Allegro Hand, and 0.76/0.75 on Shadow Hand. Gradient dot products exhibit greater variation in magnitude and across seeds. Although gradient alignment does not establish critic accuracy or agreement on auxiliary-rollout states, these findings suggest that the shared critic provides consistent local action-improvement directions despite imperfect imitation.

\section{Limitations and Future Work}
\label{sec:limitations}

Our experiments compare complete implementations of \echomethod{} and \sdpgmethod{}, without isolating the contribution of auxiliary experience from differences in architectures and training configurations. A matched rendered-only FastTD3 baseline would help quantify this contribution. Next, inverse distillation introduces approximation error because the state-based imitator may not reproduce the visual policy across all states encountered during auxiliary rollouts. While our diagnostics show positive alignment of critic action gradients despite imperfect imitation, they do not establish critic accuracy or accurate tracking on auxiliary-rollout states.

Looking ahead, \echomethod{} offers several promising research directions. First, by directly optimizing the observation-based policy rather than distilling a privileged teacher, Echo may help reduce the imitation gap arising from information asymmetry in conventional teacher--student learning. Second, the framework could extend to richer sensory modalities, such as high-resolution tactile sensing with DIGIT sensors, where simulating observations and evaluating sensory policies can be computationally expensive. Finally, decoupling experience collection from policy inference may enable scalable direct RL for larger vision-language-action (VLA) and world-action model (WAM) architectures, whose inference costs make conventional large-scale online training increasingly expensive. These directions remain to be investigated experimentally.

\section{Conclusion}

We presented a preliminary study of \echomethod{}, a framework for scaling direct visual reinforcement learning through large-scale off-policy experience. At its core, \emph{inverse distillation} trains a state-based auxiliary policy to imitate the evolving visual policy, enabling inexpensive, unrendered experience collection for off-policy value learning. A shared privileged critic leverages this experience to directly improve the visual policy through first-order gradients, decoupling the scale of experience collection from the costs of visual rendering and inference.

Our preliminary experiments show average reductions in time to reference returns of 60\% across the three locomotion tasks and 56\% across both hand tasks, with 62\% and 12\% higher peak mean returns, respectively, compared with \sdpgmethod{}. Both hands also achieve higher mean returns at matched simulator-transition budgets. These findings provide initial evidence for scaling off-policy experience without proportionally increasing visual computation. We hope that this approach offers a promising alternative to reliably scale direct visual reinforcement learning.

\clearpage
\appendix
\section{Privileged Critic Interpretation}
\label{app:critic-interpretation}
\paragraph{Interpreting Privileged Critic Gradients.}
To interpret how a privileged critic can guide an observation-based policy, consider the observation-level value function induced by the fitted critic on the rendered replay distribution:
\begin{equation}
\bar Q_{\phi_1}(o,a)
=
\mathbb E_{s\mid o}
\left[
Q_{\phi}(s,a)
\right].
\end{equation}

Holding the replay distribution and critic fixed, and assuming differentiation and conditional expectation can be interchanged, we obtain
\begin{equation}
\nabla_a\bar Q_{\phi}(o,a)
=
\mathbb E_{s\mid o}
\left[
\nabla_a Q_{\phi}(s,a)
\right].
\label{eq:state-observation-gradient}
\end{equation}

Thus, averaging critic gradients over paired state--observation samples provides an estimate of the gradient of the induced observation-level objective. This explains how privileged state information can guide an observation-based policy without becoming a deployment input. The interpretation does not imply that the fitted critic is accurate or that its gradients are unbiased estimates of the true return gradient.

\paragraph{Approximation from Auxiliary Continuation Actions.}
The state-based imitator approximates the evolving visual policy but need not reproduce its actions exactly. Consequently, TD targets constructed using $\mu_{\bar\psi}(s')$ may differ from targets obtained using the visual policy's continuation actions. In particular, the imitator is supervised on rendered visual-policy experience, whereas critic replay also contains states visited by the auxiliary policy. Low imitation error on nominal states therefore does not guarantee accurate continuation actions throughout replay.

In an idealized memoryless setting, a uniform bound on continuation-action error, together with suitable regularity assumptions, can bound the resulting value discrepancy. However, the inverse distillation objective alone does not guarantee such a bound over the full replay distribution.

\section{Gradient-Alignment Diagnostic}
\label{app:gradient-alignment}

We measure critic action-gradient alignment every 1000 visual actor updates, immediately before the corresponding optimizer step. Diagnostics use the rendered replay minibatch sampled for that update, with critic and policy parameters held fixed during measurement.

For a paired privileged state and visual observation $(s,o)$, we define
\begin{equation}
\begin{aligned}
g_v &=
\left.\nabla_a Q_\phi(s,a)\right|_{a=\pi_\theta(o)},\\
g_u &=
\left.\nabla_a Q_\phi(s,a)\right|_{a=\mu(s)},
\end{aligned}
\label{eq:diagnostic-gradients}
\end{equation}
where $\mu$ denotes either the online or EMA state-based imitator and $Q_\phi$ is the clipped twin-critic objective defined in Eq.~\eqref{eq:clipped-actor-q}.

Gradients are computed independently with respect to action tensors in float32, without differentiating through either policy. We record minibatch means of the gradient dot product $g_v^\top g_u$, cosine similarity, and both gradient norms, together with the fraction of valid samples and the fraction exhibiting positive dot products. Cosine similarity is computed only when both gradient norms exceed $10^{-8}$. All 536 diagnostic records have a valid-sample fraction of one.

For Figure~\ref{fig:diagnostics}, each seed's logged minibatch statistics are linearly interpolated onto 100 equally spaced training-time points within the common observed interval for each task. Shaded bands indicate sample standard deviation across three seeds, rather than confidence intervals over individual states.

Online and EMA gradient measurements are obtained from the same EMA-enabled training runs. Their similarity therefore does not isolate the effect of EMA on learning performance. Furthermore, these diagnostics characterize local gradients of the fitted critic on rendered replay samples; they do not establish critic accuracy or gradient alignment on auxiliary-rollout states.

\section{Hyperparameters}
\label{app:implementation}
The hyperparameters and training details are provided in Table \ref{tab:echo-hyperparameters}.


\begin{table}[ht]
\centering
\caption{\textbf{\echomethod{} training hyperparameters.}}
\label{tab:echo-hyperparameters}
\small
\setlength{\tabcolsep}{8pt}
\renewcommand{\arraystretch}{1.1}
\begin{tabular}{ll}
\toprule
\textbf{Hyperparameter} & \textbf{Value} \\
\midrule
\multicolumn{2}{l}{\textit{Simulation and observations}} \\
Total parallel environments & 4096 \\
Rendered environments & 64 \\
Auxiliary environments & 4032 \\
Image resolution & $84\times84$ RGB \\
Replay history per environment & 512 transitions \\
\midrule
\multicolumn{2}{l}{\textit{Off-policy critic learning}} \\
RL algorithm & FastTD3 \\
Critic architecture & Twin categorical critics \\
Distributional return atoms & 101 \\
Critic batch size & 32768 \\
Critic updates per collection step & 2 \\
Target-policy smoothing & Clipped Gaussian \\
\midrule
\multicolumn{2}{l}{\textit{Visual policy learning}} \\
Visual actor batch size & 256 \\
Actor update frequency & Every 2 critic updates \\
Visual target actor & None \\
\midrule
\multicolumn{2}{l}{\textit{Inverse distillation}} \\
Auxiliary imitation objective & Action MSE \\
Auxiliary updates per collection step & 1 \\
Auxiliary collection policy & Online imitator \\
TD continuation policy & EMA imitator \\
Auxiliary EMA coefficient $\tau_\mu$ & 0.1 \\
\midrule
\multicolumn{2}{l}{\textit{Evaluation}} \\
Training seeds & 0, 1, 2 \\
Evaluation episodes & 64 \\
Evaluation policy & Deterministic visual actor \\
GPU & NVIDIA H100 \\
\bottomrule
\end{tabular}
\end{table}

\section{Aggregate Performance Comparisons}
\label{app:aggregate-performance}
For each task, the reference is the \sdpgmethod{} three-run mean return at the shared budget. After linear interpolation over the common observed interval, $T$ is the first crossing time of this reference and $P$ the peak of the three-run mean curve. We report time reduction $100(1-T_E/T_S)\%$ and return improvement $100(P_E/P_S-1)\%$, where $E$ and $S$ denote \echomethod{} and \sdpgmethod{}, and average unrounded percentages with equal task weights.

\begin{center}
\begin{tabular}{lrr}
\toprule
Task or task family & Time reduction & Peak mean-return gain \\
\midrule
Hopper & 64\% & 23\% \\
Walker & 25\% & 27\% \\
Ant & 89\% & 136\% \\
Allegro Hand & 47\% & 16\% \\
Shadow Hand & 65\% & 9\% \\
\midrule
\textbf{Locomotion average} & \textbf{60\%} & \textbf{62\%} \\
\textbf{Manipulation average} & \textbf{56\%} & \textbf{12\%} \\
All five tasks & 58\% & 42\% \\
\bottomrule
\end{tabular}
\end{center}

\end{document}